\documentclass[11pt]{article}
\usepackage[a4paper]{geometry}
\usepackage{authblk}
\usepackage{clic2017}
\usepackage{times}
\usepackage{url}
\usepackage{latexsym}
\usepackage{color}

\usepackage{amsmath,amsfonts,amssymb}
\usepackage{graphicx}
\usepackage[multiple]{footmisc}
\usepackage[T1]{fontenc}
\usepackage[utf8]{inputenc}
\title{Almawave-SLU: A new dataset for SLU in Italian}

\author{\textbf{Valentina Bellomaria}}
\author{\textbf{Giuseppe Castellucci}}
\author{\textbf{Andrea Favalli}}
\author{\textbf{Raniero Romagnoli}}
\affil{Language Technology Lab \\Almawave srl \\\textit{[first name initial].[last name]}@almawave.it}

\date{}

\begin{document}
\maketitle
\begin{abstract}
  \textbf{English.}  
  
The widespread use of conversational and question answering systems made it necessary to improve the performances of speaker intent detection and understanding of related semantic slots, i.e., Spoken Language Understanding (SLU). Often, these tasks are approached with supervised learning methods, which needs considerable labeled datasets. This paper presents the first Italian dataset for SLU. It is derived through a semi-automatic procedure and is used as a benchmark of various open source and commercial systems.
\end{abstract}

\begin{abstract-alt}
 \textrm{\bf{Italiano.}} 

La diffusione di interfacce conversazionali e sistemi di question answering ha reso necessario migliorare le prestazioni degli algoritmi in grado riconoscere l'intento del parlante e la comprensione dei suoi argomenti semantici. Questi task vengono generalmente implementati con tecniche di apprendimento supervisionato, che necessitano di considerevoli quantità di dati etichettati. In questo paper, viene presentato il primo dataset per la lingua italiana per questi task, realizzato tramite una procedura semi-automatica e utilizzato come benchmarck di vari sistemi, sia open source che commerciali.
\end{abstract-alt}

\section{Introduction}
\label{sec:intro}

\begin{figure}[t]
  \centering
    \includegraphics[width=0.46\textwidth]{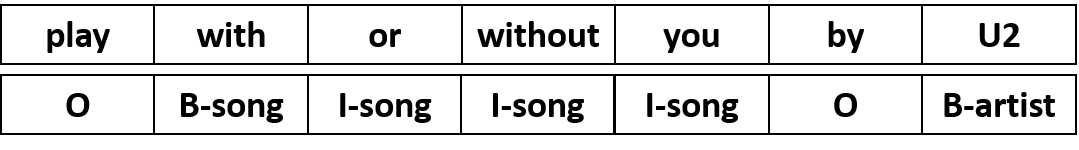}
  \caption{An example of Slot Filling in IOB format for a sentence with intent \textit{PlayMusic}.}
  \label{fig:slu-example}
\end{figure}

Conversational interfaces, e.g., Google's Home or Amazon's Alexa, are becoming pervasive in daily life. As an important part of any conversation, language understanding aims at extracting the meaning a partner is trying to convey. Spoken Language Understanding (SLU) plays a critical role in such a scenario. Generally speaking, in SLU a spoken utterance is first transcribed, then semantic information is extracted. Language understanding, i.e., extracting a semantic ``frame'' from a transcribed user utterance, typically involves: i) Intent Detection (ID) and ii) Slot Filling (SF) \cite{tur2011}. The former makes the classification of a user utterance into an intent, i.e., the purpose of the user. The latter finds what are the ``arguments'' of such intent. As an example, let us consider Figure \ref{fig:slu-example}, where the user asks for playing a song (\texttt{Intent=PlayMuysic}) (\textit{with or without you}, \texttt{Slot=song}) of an artist (\textit{U2}, \texttt{Slot=artist}).

Usually, supervised learning methods are adopted for SLU. Their efficacy strongly depends on the availability of labeled data. %Thus, data labeling is an indispensable task but it is also time-consuming.
There are various approaches to the production of labeled data, depending on the complexity of the problem, on the characteristics of the data, and on the available resources (e.g., annotators, time and budget). When the reuse existing public data is not feasible, manual labeling should be accomplished, eventually by automating part of the labeling process.

In this work, we present the first public dataset for the Italian language for SLU. It is generated through a semi-automatic procedure from an existing English dataset annotated with intents and slots. We have translated the sentences into Italian and reported the annotations based on a token span algorithm.
%and performed an automatic replacement of some entities on the slots.
Then, the translation, spans and consistency of the entities in Italian have been manually validated. Finally, the dataset is used as benchmark for NLU systems. In particular, we will compare a recent state-of-the-art (SOTA) approach \cite{ernie} with Rasa \cite{rasa} taken from the open source world, IBM Watson Assistant \cite{watson}, Google DialogFlow \cite{dialogflow} and, finally, Microsoft LUIS \cite{msluis}, some commercial solutions in use.

Following, in section \ref{sec:related} related works will be discussed; section \ref{sec:production} will discuss the dataset generation. Section \ref{sec:experiments} will present the experiments. Finally, section \ref{sec:conclusion} will derive the conclusions.

\section{Related Work}
\label{sec:related}

SLU has been addressed in the Natural Language Processing community mainly in the English language. 
A well-known dataset used to demonstrate and benchmark various NLU algorithms is Airline Travel Information System (ATIS) \cite{atis1990} dataset, which consists of spoken queries on flight related information. In \cite{braun-EtAl:2017:SIGDIAL} were presented three dataset for Intent classification task. \textit{AskUbuntu Corpus} and \textit{Web Application Corpus} were extracted from StackExchange and the third one, i.e., \textit{Chatbot Corpus}, was derived from a Telegram chatbot. The newer multi-intent dataset SNIPS \cite{snips2018} is the starting point for the work presented in this paper.
An alternative approach to manual or semi-automatic labeling is the one proposed by the data scientists of the Snorkel project with Snorkel Drybell \cite{2018arXiv181200417B} that aims at the automate the labeling through the use of data programming. Other works have explored the possibility of creating datasets in a language starting from datasets of other languages, such as \cite{Jabaian2010InvestigatingMA} and \cite{inproceedings}.

\section{Almawave-SLU: A new dataset for Italian SLU}
\label{sec:production}

%To  the  best  of  our  knowledge,  there  is  no  annotated dataset for SLU in Italian.
We derived the new dataset \footnote{The dataset will be available for download} starting from the SNIPS dataset \cite{snips2018}, which is in English. It contains $14,484$ annotated examples\footnote{There are $13084$, $700$ and $700$ for training, validation and test, respectively.} with respect to $7$ intents and $39$ slots. In table \ref{tab:snipsexamples} an excerpt of the dataset is shown. We started from this dataset as: i) it contains a reasonable amount of examples; ii) it is multi-domain; iii) we believe it could represent a more realistic setting in today's voice assistants scenario.

\begin{table*}[!t]
\footnotesize
\centering
\begin{tabular}{|l|l|}
\hline 
\texttt{AddToPlaylist} & Add the song virales de siempre by the cary brothers to my gym playlist. \\ \hline
\texttt{BookRestaurant} & I want to book a top-rated brasserie for 7 people.  \\ \hline
\texttt{GetWeather} & What kind of weather will be in Ukraine one minute from now?  \\ \hline
\texttt{PlayMusic} & Play Subconscious Lobotomy from Jennifer Paull.  \\ \hline
\texttt{RateBook} & Rate The children of Niobe 1 out of 6 points. \\ \hline
\texttt{SearchCreativeWork}   & Looking for a creative work called Plant Ecology \\ \hline
\texttt{SearchScreeningEvent}  & Is Bartok the Magnificent playing at seven AM?  \\ \hline

\end{tabular}
\caption{Examples from the SNIPS dataset. The first column indicates the intent, the second columns contains an example.}
\label{tab:snipsexamples}
\end{table*}

We performed a semi-automatic procedure consisting of two phases: an automatic translation with contextual alignment of intents and slots; a manual validation of the translations and annotations.
%In the following, details about the two steps are provided, as well as more information about the difficulties we found in producing the Italian version of the dataset.
The resulting dataset, i.e., \texttt{Almawave-SLU}, has fewer training examples, a total of $7,142$ and the same number of validation and test examples of the original dataset. Again, $7$ intents and $39$ slots have been annotated.
Table \ref{tab:datastats} shows the distribution of examples for each intent.

\begin{table}[b]
\footnotesize
\centering
\begin{tabular}{|l|c|c|c|c|}
\hline
& \textbf{\scriptsize{Train}} & \textbf{ \scriptsize{Train-R}} & \textbf{\scriptsize{Valid}} & \textbf{\scriptsize{Test}}   \\ \hline
\texttt{\scriptsize{AddToPlayList}}& $744$ & $185$ & $100$ & $124$  \\ \hline
\texttt{\scriptsize{BookRestaurant}} & $967$ & $250$ & $100$ & $92$   \\ \hline
\texttt{\scriptsize{GetWeather}} & $791$ & $195$& $100$ & $104$  \\ \hline
\texttt{\scriptsize{PlayMusic}} & $972$ & $240$ & $100$ & $86$ \\ \hline
\texttt{\scriptsize{RateBook}} & $765$ & $181$ & $100$ & $80$ \\ \hline
\texttt{\scriptsize{SearchCreativeWork}} & $752$ & $172$ & $100$ & $107$ \\ \hline
\texttt{\scriptsize{SearchScreeningEvent}} & $751$ & $202$ & $100$ & $107$ \\ \hline
\end{tabular}
\caption{\texttt{Almawave-SLU} Datasets statistics. Train-R is the reducted training set.}
\label{tab:datastats}
\end{table}

\subsection{Translation and Annotation}
\label{sec:translation}

In a first phase, we translated each English example in Italian by using the Translator Text API: part of the Microsoft Azure Cognitive Services. %\footnote{\url{https://docs.microsoft.com/en-us/azure/cognitive-services/translator/translator-info-overview}}.
In order to create a more valuable resource in Italian, we also performed an automatic substitution of the names of movies, movie theatres, books, restaurants and of the locations with some Italian counterpart. First, we collected from the Web a set $E$ of about $20,000$ Italian versions of such entities; then, we substituted each entity in the sentences of the dataset with one randomly chosen from $E$.

After the translation, an automatic annotation was performed. The intent associated with the English sentence has been copied to its Italian counterpart. Slots have been transferred by aligning the source and target tokens\footnote{The alignment was provided by the Translator API.} and by copying the corresponding slot annotation.
In  case  of  exceptions, e.g., multiple alignments on the same token or missing alignment, we left the token without annotation.
    
\subsection{Human Revision}
\label{sec:human}

In a second phase, the dataset was splitted into $6$ different sets, each containing about $1,190$ sentences. Each set was assigned to $2$ annotators\footnote{A total of $6$ annotators were available.}, and each was asked to review the translation from English to Italian and the reliability of the automatic annotation. The guideline was to consider a valid annotation when both the alignment and the semantic slots were correct. Moreover, also a semantic consistency check was performed: e.g., served dish and restaurant type or city and region or song and singer.
%Since the construction of linguistic resources implies subjectivity in the identification of linguistic phenomena and in the assignment of a category to the issue in analysis, to guarantee the reliability of the annotated resource and the reproducibility of the annotation process, we considered that at least two annotators had to be in place on the same annotation.
The $2$ annotators have been used to cross-check the annotations, in order to provide more reliable revisions. When the $2$ annotators disagreed, the annotations have been validated by a third different annotator.

During the validation phase some interesting phenomena emerged. \footnote{Some inconsistencies were in the original dataset} For example, there have been cases of inconsistency between the restaurant name and the type of served dish when the name of the restaurant mentioned the kind of food served, e.g., %"Book a reservation at "Pizza Point" for noodle,
\textit{"Prenota un tavolo da Pizza Party per mangiare noodles"}.
There were also wrong associations between the type of restaurant and service requested, e.g, %"Make a reservation in the pool area for 4 people at a truck restaurant" 
\textit{"Prenota nell'area piscina per 4 persone in un camion-ristorante"}. A truck restaurant is actually a van equipped for fast-food in the street.
Again, among the cases of unlikely associations resulting from automatic replacement, the inconsistency between temperatures and cities is mentioned, in cases like "snow in the Sahara". Another type of problem occured when the same slot was used to identify very different objects. For example, for the intent \textit{SearchCreativeWork}, the slot \textit{object\_name} was used for paintings, games, movies, etc...
%We can observe and analyze a couple of examples for this intent: \textit{Can you find me the work, The Curse of Oak Island ?} and \textit{Can you find me, Hey Man ?}.
%The first example contains \textit{The Curse of Oak Island}, that is a television series and the second refers to \textit{Hey Man} that is a music album, but both are labeled as \textit{object\_name}, where the \textit{object\_type} are different and not specified. 
In all these cases, the annotators were asked to correct the sentences and the annotations, accordingly. Again, in the case of \textit{BookRestaurant} intent a manual revision was made when in the same sentence the city and state coexist: to make the data more relevant to the Italian language, the region relative to the city is changed, e.g, \textit{"I need a table for 5 at a highly rated gastropub in Saint Paul, MN"} is translated and adapted for Italian in \textit{"Vorrei prenotare un tavolo per 5 in un gastropub molto apprezzato a Biella, Piemonte"}.
%In the cases of \textit{PlayMusic} and \textit{AddToPlaylist} intent there have been many translation problems since the titles were translated when they shouldn't have been, e.g \textit{"Put What Color Is Your Sky by Alana Davis on the stereo"} was transated as \textit{"Metti quello che colore è il tuo cielo di Alana Davis allo stereo."} .
\subsection{Automatic Translation Analysis}
\label{sec:analysis}

In many cases, machine translation lacked context awareness: this isn't an easy task due to phenomena as polysemy, homonymy, metaphors and idioms. There can be problems of lexical ambiguities when a word has more than one meaning and can produce wrong interpretations. For example, the verb "to play" can mean ‘‘spend time doing enjoyable things'', such as ‘‘using toys and taking part in games'', ‘‘perform music'' or ‘‘perform the part of a character''.

Human intervention occurred to maintain the meaning of the text dependent on cultural and situational contexts. Different translation errors were modified by the annotators. For example, the automatic translation of the sentence \textit{Play Have You Met Miss Jones by Nicole from Google Music.} was \textit{Gioca hai incontrato Miss Jones di Nicole da Google Music.}, but the correct Italian version is \textit{Riproduci Have You Met Miss Jones di Nicole da Google Music.}. In this case the wrong translation of the verb \textit{play} causes a meaningless sentence.

Often, translation errors are due to presence of prepositions, that have the same function in Italian as they do in English. Unfortunately, these cannot be directly translated. Each preposition is represented by a group of related senses, some of which are very close and similar while others are rather weak and distant. For example, the Italian preposition ‘‘di'' can have six different English counterparts – of, by, about, from, at, and than. 
For example, in the SNIPS dataset the sentence \textit{I need a table for 2 on feb. 18 at Main Deli Steak House} was translated as \textit{Ho bisogno di un tavolo per 2 su Feb. 18 presso Main Deli Steak House}. Here, the translation of ‘‘on'' is wrong: the correct Italian version should translate it as ‘‘il''. Another example with wrong preposition translation is the sentence \textit{‘‘What will the weather be one month from now in Chad ?'}, the automatic translation of ‘‘one month from now'' is ‘‘un mese da ora'' but the correct translation is ‘‘tra un mese''. 

Common errors were in the translation of temporal expression, that are different between Italian and English. For example the translation of the sentence \textit{‘‘Book a table in Fiji for zero a.m''} was \textit{‘‘Prenotare un tavolo in Fiji per zero a.m}" but in Italian ‘‘zero a.m'' is ‘‘mezzanotte''.

Other errors were specific of some intents, as they tend to have more slangs.
For example, the translation of \textit{GetWeather}'s sentences was problematic because the main verb is often misinterpreted, while in the sentences related to the intent \textit{BookRestaurant} a frequent failure occurred on the interpretation of prepositions. For example, the sentence \textit{‘‘Will it get chilly in North Creek Forest?''} was translated as ‘‘Otterrà freddo in North Creek Forest?'', while the correct translation is ‘‘Farà freddo a North CreekForest?''. In this case, the system misinterpreted the context, assigning to ``get'' the wrong meaning.

\section{Benchmarking SLU Systems}
\label{sec:experiments}

Nowadays, there are many human-machine interaction platforms, commercial and open source. %Some, such as Facebook’s Wit.ai, AmazonLex, Microsoft’s LUIS, IBM’s Watson Assistant and Google’s Dialogflow are commercial. Others, such as Rasa are open source. 
Machine learning algorithms enables these systems to understand natural language utterances, match them to intents, and extract structured data. %The Almawave-SLU dataset could be used for evaluating those systems. We decided to use the dataset to test several systems: Rasa, LUIS, Watson Assistant, Dialogflow and Bert-Joint that represent the SOTA.
We decided to use the Almawave-SLU dataset with the following SLU systems.

\subsection{SLU Systems}
\label{sec:slu-systems}

\paragraph{RASA.} RASA \cite{rasa} is an open source alternative to popular NLP tools for the classification of intentions and the extraction of entities. Rasa contains a set of high-level APIs to produce a language parser through the use of NLP and ML libraries, via the configuration of the pipeline and embeddings. It seems to be very fast to train, does not require great computing power and, despite this, it seems to get excellent results. % It is also configurable and it allows to choose custom model pipelines, but it also provides two standard ones, the first uses pre-trained word embeddings, while the latter does not.

\paragraph{LUIS.} Language Understanding service \cite{msluis} allows the construction of applications that can receive input in natural language and extract the meaning from it through the use of Machine Learning algorithms. %Provide APIs for uploading training examples and for calculating predictions. LUIS also provides a user-friendly user interface for the use of all features by less experienced users. 
LUIS was chosen as it provides also an easy-to-use graphical interface dedicated to less experienced users. For this system the computation is done completely remotely and no configurations are necessary.

\paragraph{Watson Assistant.} 
%IBM's Watson Assistant \cite{watson} is a white label cloud service that allows software developers to embed a virtual assistant, that use Watson AI machine learning and natural language understanding, in their software. Access to Watson AI is delivered through the IBM Cloud. Watson Assistant enable customers to isolate and protect the information their assistant gathers through user interaction in a private cloud. Watson Assistant was chosen because it is a product that has a long tradition in this task, it has been built for an industrial market and it runs on the IBM Cloud.
IBM's Watson Assistant \cite{watson} is a white label cloud service that allows software developers to embed a virtual assistant, that use Watson AI machine learning and NLU, in their software. Watson Assistant allows customers to protect information gathered through user interaction in a private cloud. It was chosen because it was conceived for an industrial market and for its long tradition in this task.

\paragraph{DialogFlow.} Dialogflow \cite{dialogflow} is a Google service to build engaging voice and text-based conversational interfaces, %such as voice apps and chatbots, 
powered by a natural language understanding (NLU) engine. %that process and understand natural language input. 
Dialogflow makes it easy to connect the bot service to a number of channels and runs on Google Cloud Platform, so it can scale to hundreds of millions of users. DialogFlow was chosen due to its wide distribution and ease of use of the interface.

\paragraph{Bert-Joint.} It is a SOTA approach to SLU adopting a joint Deep Learning architecture in an attention-based recurrent frameworks \cite{ernie}. It exploits the successful Bidirectional Encoder Representations from Transformers (BERT) model to pre-train language representations. In \cite{ernie}, the authors extend the BERT model in order to perform the two tasks of ID and SF jointly. In particular, two classifiers are trained jointly on top of the BERT representations by means of a specific loss function.

\subsection{Experimental Setup}
\label{sect:expsetup}

\texttt{Almawave-SLU} has been used for training and evaluation of Rasa, Luis, Watson Assistant, DialogFlow and Bert-Joint.
Another evalution is made on $3$ different training datasets, i.e Train-R, of reduced dimensions with respect to the Almawave-SLU, each about $1,400$ sentences equally distributed on intent.

\begin{table*}[!ht]
\footnotesize
\centering
\begin{tabular}{|l|c|c|c|c|c|c|}
\hline 
 \hline
& \multicolumn{3}{|c|}{\textbf{Eval-1 with Train set}} & \multicolumn{3}{|c|}{\textbf{ Eval-2 with  Train-R set}} \\
System & Intent &  Slot & Sentence & Intent &  Slot & Sentence \\ \hline
\textbf{Rasa} & $96.42$ & $85.40$ & $65.76$ & $93.84$ & $78.58$ & $52.25$\\ \hline
\textbf{LUIS} & $95.99$ & $79.47$ & $50.57$& $94.46$ & $72.51$ & $35.53$ \\ \hline
\textbf{Watson Assistant} & $96.56$ & - & - & $95.03$ & - & - \\ \hline
\textbf{Dialogflow} & $95.56$ & $74.62$ & $46.16$ & $93.60$ & $65.23$ & $36.68$ \\ \hline
\textbf{Bert-Joint} & $\textbf{97.6}$ & $\textbf{90.0}$ & $\textbf{77.1}$ & $\textbf{96.13}$ & $\textbf{83.04}$ & $\textbf{65.23}$ \\ \hline

\end{tabular}
\caption{Overall scores for Intent and Slot}
\label{tab:overallScores}
\end{table*}

The train/validation/test split used for the evaluations is $5,742$ ($1,400$ for Train-R), $700$ and $700$, respectively.
Regarding Rasa, we used version $1.0.7$, and we adopted the standard ‘‘supervised embeddings'' pipeline, since it is recommended in the official documentation. This pipeline consists of a 
\textit{WhiteSpaceTokenizer}, that was modified to avoid the filter of punctuation tokens, a \textit{Regex Featurizer}, a \textit{Conditional Random Field} to extract entities, a \textit{Bag-of-words Featurizer} and an \textit{Intent Classifier}.
LUIS was tested against the api v$2.0$, and the loading of data to train the system with LUIS APP VERSION $0.1$.
%The data were converted into JSON format for each platform according to the specific requirements.
%Furthermore, the Appendix provides an input format example for each system.
Unfortunately Watson Assistant supports only English models for the annotations of contextual entities, i.e, slots; therefore, we have only measured the intents \footnote{Refer to \textit{Table 3. Entity feature support details} at \url{ https://cloud.ibm.com/docs/services/assistant?topic=assistant-language-support}}.
Regarding DialogFlow, a ‘‘Standard'' (free) utility has been created with API version 2; the python library ‘‘dialogflow'' has been used for the predictions.  \footnote{\url{https://cloud.google.com/dialogflow/docs/reference/rest/v2/projects.agent.intents#Part}}. %DialogFlow allows to choice between pure ML mode and hybrid rule-based and ML mode. We chosen ML mode.
We changed only the setting ‘‘match mode'' to ‘‘ML only''.
Regarding the BERT-Joint system, a pre-trained BERT model is adopted, which is available on the BERT authors website\footnote{\url{https://storage.googleapis.com/bert\_models/2018\_11\_23/multi\_cased\_L-12\_H-768\_A-12.zip}}. This model is composed of $12$-layer and the size of the hidden state is $768$. The multi-head self-attention is composed of $12$ heads for a total of $110$M parameters.
As suggested in \cite{ernie}, we adopted a dropout strategy applied to the final hidden states before the intent/slot classifiers.
We tuned the following hyper-parameters over the validation set: (i) number of epochs among ($5$, $10$, $20$, $50$); (ii) Dropout keep probability among ($0.5$, $0.7$ and $0.9$). We adopted the Adam optimizer \cite{kingma2014} with parameters $\beta_1=0.9$, $\beta_2=0.999$, L$2$ weight decay $0.01$ and learning rate $2\text{e-}5$ over batches of size $64$.

\subsection{Experimental Results}
\label{sec:results}

In table \ref{tab:overallScores} the performances of the systems are shown. The SF performance is the F1 while the ID and Sentence performances are measured with the accuracy.
%We used the conlleval script to evaluate the systems . This script can be used with IOB2 or IOBES tagging scheme. For the LUIS and Rasa systems that use a different scheme, these has been converted to IOB format.
We also show an evaluation carried out with models trained on three different split of reduced size derived from the whole dataset. The reported value is the average of measurements obtained separately on the entire test dataset.

Regarding the ID task, all models are performing similarly, but Bert-Joint F1 score is little higher than others. For SF task, notice that there are significant differences between LUIS, DialogFlow and Rasa performances. 

%The value of sentence accuracy must be interpreted considering that the probability of correctly classifying both the intent and all the slots of a sentence using a random algorithm is close to zero. Assuming there are m> 0 intents and n> 0 entities with the same number of occurrences and that each sentence has k> 0 tokens, this probability is similar to ((1 / n) ^ k)) * (1 / m).
%Finally, Bert-Joint achieved the top sentence accuracy ($77.1$) on joint classification. 
Finally, Bert-Joint achieved the top score on joint classification, in the assessments with the two different sizes of the dataset. The adaptation of nominal entities in Italian may have amplified the problem for the other models.

\section{Conclusion}
\label{sec:conclusion}

The contributions of this work are two-fold: first, we presented and released the first SLU dataset in Italian: (\texttt{Almawave-SLU}), composed of $7,142$ sentences annotated with respect to intents and slots, almost equally distributed on the $7$ different intents. The effort spent on the construction of this new resource, according to the semi-automatic procedure described, is about 24 FTE \footnote{Full Time Equivalent}, with an average production of about $300$ examples per day. We consider this effort lower than typical efforts to create linguistic resources from scratch.
%As we all know, labeling training data is one of the most costly bottlenecks in developing machine learning-based applications. The dataset represents a starting point for the evaluation and comparison of task-oriented systems.

Second, we compared some of the most popular NLU services with this data. The results show they all have similar features and performances. However, compared to another specific architecture for SLU, i.e., Bert-Joint, they perform worse. It was expected and it demonstrates the Almawave-SLU can be a valuable dataset to train and test SLU systems on the Italian language.
In future, we hope to continuously improve the data and to extend the dataset.

\section{Acknowledgment}
\label{sec:ack}

The authors would like to thank to David Alessandrini, Silvana De Benedictis, Raffaele Mazzocca, Roberto Pellegrini and Federico Wolenski for the support in the annotation, revision and evaluation phases.

%\section{Appendix}
%\label{sec:appendix}

%\textbf{RASA:}
%   {
%    "text": "Prenota Ristorante Pizzeria Kiss me in Molise per mezzogiorno",
%    "intent": "BookRestaurant",
%    "entities": [
%     {
%      "start": 8,
%      "end": 35,
%      "value": "Ristorante Pizzeria Kiss me",
%      "entity": "restaurant_name"
%     },
%     {
%      "start": 39,
%      "end": 45,
%      "value": "Molise",
%      "entity": "state"
%     },
%     {
%      "start": 50,
%      "end": 61,
%      "value": "mezzogiorno",
%      "entity": "timeRange"
%     }
%    ]
%   }
%\textbf{LUIS:}
% {
%  "text": "Prenota Ristorante Pizzeria Kiss me in Molise per mezzogiorno",
%  "intentName": "BookRestaurant",
%  "entityLabels": [
%   {
%    "entityName": "restaurant_name",
%    "startCharIndex": 8,
%    "endCharIndex": 34
%   },
%   {
%    "entityName": "state",
%    "startCharIndex": 39,
%    "endCharIndex": 44
%   },
%   {
%    "entityName": "timeRange",
%    "startCharIndex": 50,
%    "endCharIndex": 60
%   }
%  ]
% }
%\textbf{BERT-JOINT:}
%{"id": 1, "sentence": "Prenota Ristorante Pizzeria Kiss me in Molise per mezzogiorno", "category": "BookRestaurant", "tokens": ["Prenota", "Ristorante", "Pizzeria", "Kiss", "me", "in", "Molise", "per", "mezzogiorno"], "labels": ["O", "B-restaurant_name", "I-restaurant_name", "I-restaurant_name", "I-restaurant_name", "O", "B-state", "O", "B-timeRange"]}

\bibliography{clic2019}
\bibliographystyle{acl}

\end{document}